\documentclass[10pt, letterpaper]{article}

\usepackage{cogsci}
\usepackage{pslatex}
\usepackage{float}
\usepackage{caption}

\usepackage{amsmath}

\usepackage{amssymb}

\usepackage{hyperref}

\usepackage{graphicx}

\usepackage{fancyvrb}
\DefineVerbatimEnvironment{Sinput}{Verbatim}{fontshape=sl}
\DefineVerbatimEnvironment{Soutput}{Verbatim}{}
\DefineVerbatimEnvironment{Scode}{Verbatim}{fontshape=sl}

\DefineVerbatimEnvironment{Code}{Verbatim}{}
\DefineVerbatimEnvironment{CodeInput}{Verbatim}{fontshape=sl}
\DefineVerbatimEnvironment{CodeOutput}{Verbatim}{}
\newenvironment{CodeChunk}{}{}

\usepackage{apacite}

\cogscifinalcopy

\usepackage{color}

\title{Generalizing meanings from partners to populations:\\
Hierarchical inference supports convention formation on networks}

\author{{\large \bf Robert D.~Hawkins \textsuperscript{1}, Noah D.~Goodman\textsuperscript{2}, Adele E.~Goldberg\textsuperscript{1}, Thomas L.~Griffiths\textsuperscript{1,3}} \\
 \textsuperscript{1}Department of Psychology, Princeton University (\texttt{\{rdhawkins,adele,tomg\}@princeton.edu}) \\
\textsuperscript{2}Department of Psychology and Computer Science, Stanford University (\texttt{ndgoodman@stanford.edu})\\ \textsuperscript{3}Department of Computer Science, Princeton University}

\begin{document}

\maketitle

\begin{abstract}
A key property of linguistic conventions is that they hold over an
entire community of speakers, allowing us to communicate efficiently
even with people we have never met before. At the same time, much of our
language use is partner-specific: we know that words may be understood
differently by different people based on our shared history. This poses
a challenge for accounts of convention formation. Exactly how do agents
make the inferential leap to community-wide expectations while
maintaining partner-specific knowledge? We propose a hierarchical
Bayesian model to explain how speakers and listeners solve this
inductive problem. To evaluate our model's predictions, we conducted an
experiment where participants played an extended natural-language
communication game with different partners in a small community. We
examine several measures of generalization and find key signatures of
both partner-specificity and community convergence that distinguish our
model from alternatives. These results suggest that partner-specificity
is not only compatible with the formation of community-wide conventions,
but may facilitate it when coupled with a powerful inductive mechanism.

\end{abstract}

To communicate successfully, speakers and listeners must share a common
system of semantic meaning in the language they are using. These
meanings are \emph{conventional} in the sense that they are sustained by
the expectations each person has about others (Hawkins et al., 2019a;
Lewis, 1969). A key property of linguistic conventions is that they hold
over an entire community of speakers, allowing us to communicate
efficiently even with people we've never met before. But exactly how do
we make the inferential leap to community-wide expectations from our
experiences with specific partners? Grounding collective convention
formation in individual cognition requires an explicit \emph{theory of
generalization} capturing how people transfer what they have learned
from one partner to the next.

One influential theory is that speakers simply ignore the identity of
different partners and update a single monolithic representation after
every interaction (Barr, 2004; Steels, 1995; Young, 2015). We call this
a \emph{complete-pooling} theory because data from each partner is
collapsed into an undifferentiated pool of evidence (Gelman \& Hill,
2006). Complete-pooling models have been remarkably successful at
predicting collective behavior on networks, but have typically been
evaluated only in settings where anonymity is enforced. For example,
Centola and Baronchelli (2015) asked how large networks of participants
coordinated on conventional names for novel faces. On each trial,
participants were paired with a random neighbor but were not informed of
that neighbor's identity, or the total number of different possible
neighbors.

While complete-pooling may be appropriate for some everyday social
interactions, such as coordinating with anonymous drivers on the
highway, it is less tenable for everyday communicative settings.
Knowledge about a partner's identity is both available and relevant for
conversation (Davidson, 1986; Eckert, 2012). Extensive evidence from
psycholinguistics has demonstrated the \emph{partner-specificity} of our
language use (Clark, 1996). Because meaning is grounded in the evolving
`common ground' shared with each partner, meanings established over a
history of interaction with one partner are not necessarily transferred
to other partners (Metzing \& Brennan, 2003; Wilkes-Gibbs \& Clark,
1992). Partner-specificity thus poses clear problems for
complete-pooling theories but can be easily explained by another simple
model, where agents maintain separate expectations about meaning for
each partner. We call this a \emph{no-pooling} model. The problem with
no-pooling is that agents are forced to start from scratch with each
partner. Community-level expectations never get off the ground.

What theory of generalization, then, can explain partner-specific
meaning but also allow conventions to spread through communities? We
propose a \emph{partial-pooling} account that offers a compromise
between these extremes. Unlike complete-pooling and no-pooling models,
we propose that beliefs about meaning have hierarchical structure. That
is, the meanings used by different partners are expected to be drawn
from a shared community-wide distribution but are also allowed to differ
from one another in systematic, partner-specific ways. This structure
provides an inductive pathway for abstract population-level expectations
to be distilled from partner-specific experience (see also Kleinschmidt
\& Jaeger, 2015; Tenenbaum, Kemp, Griffiths, \& Goodman, 2011).

We begin by formalizing this account in a probabilistic model of
communication and presenting several simulations of listener and speaker
behavior within and across partners. Next, we test the qualitative
predictions of this model in a behavioral experiment. Participants were
paired for a series of extended reference games with each neighbor in
small networks. Our results showed signatures of partner-specific
convention formation within pairs, but also gradual generalization of
these local conventions across subsequent partners as the network
converged. Taken together, these results suggest that local
partner-specific learning is not only compatible with global convention
formation but may facilitate it when coupled with a powerful
hierarchical inference mechanism.

\hypertarget{a-hierarchical-bayesian-model-of-convention}{%
\section{A hierarchical Bayesian model of
convention}\label{a-hierarchical-bayesian-model-of-convention}}

\begin{CodeChunk}
\begin{figure}[b!]

{\centering \includegraphics[width=200px]{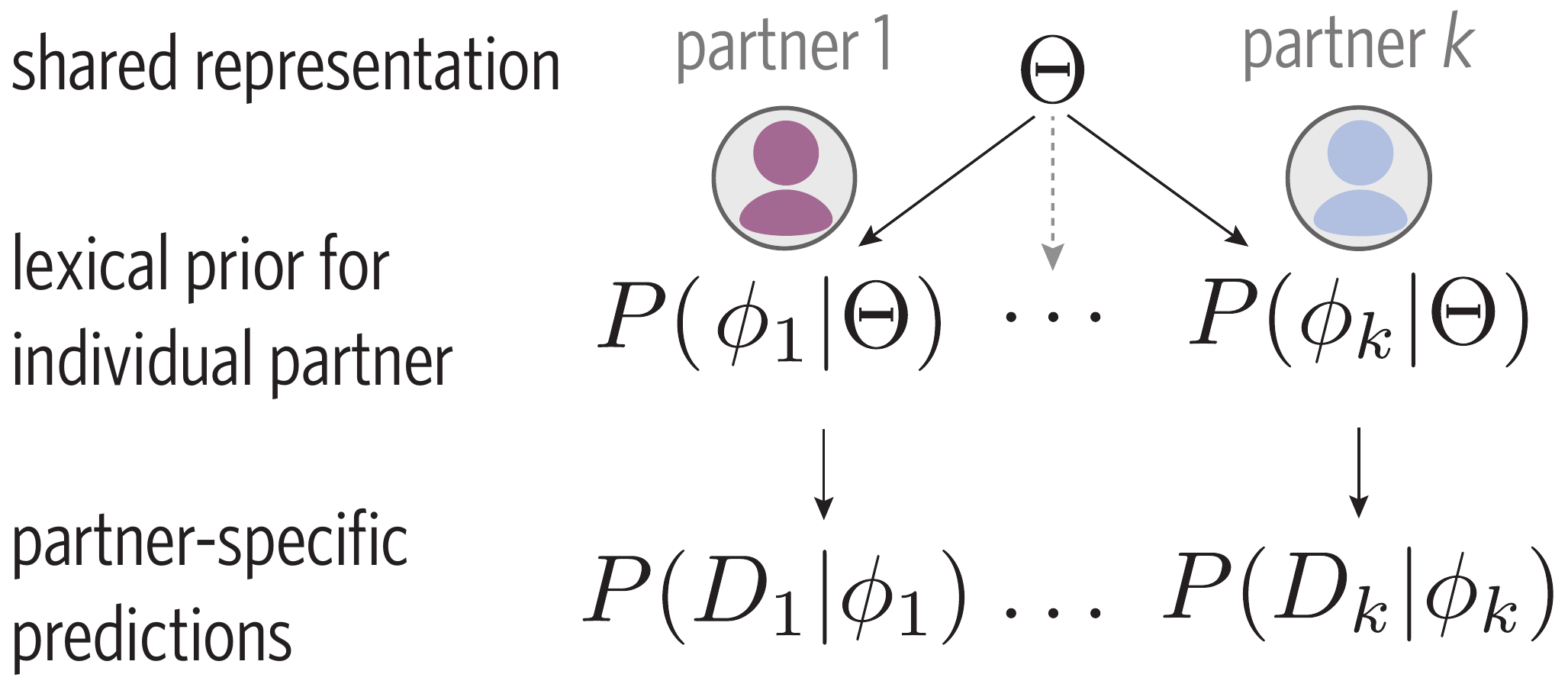} 

}

\caption{\label{fig:task1model} Schematic of hierachical Bayesian model.}\label{fig:model_schematic}
\end{figure}
\end{CodeChunk}

In this section, we provide an explicit computational account of the
cognitive mechanisms supporting the balance between community-level
stability and partner-specific flexibility. Specifically, we show how
the Bayesian model of dyadic convention formation proposed by Hawkins,
Frank, and Goodman (2017) can be extended with a principled mechanism
for generalization across multiple partners. This model begins with the
idea that knowledge about linguistic meaning can be represented
probabilistically: agents have uncertainty about the form-meaning
mappings their current partner is using (Bergen, Levy, \& Goodman,
2016). In our hierarchical model, this uncertainty is represented at
multiple levels of abstraction.

At the highest level of the hierarchy is \emph{community-level}
uncertainty \(P(\Theta)\), where \(\Theta\) represents an abstract
``overhypothesis'' about the overall distribution of possible partners.
\(\Theta\) then parameterizes the agent's \emph{partner-specific}
uncertainty \(P(\phi_{k} | \Theta)\), where \(\phi_k\) represents the
specific system of meaning used by partner \(k\) (see Fig.
\ref{fig:task1model}). Given observations \(D_k\) from interactions with
partner \(k\), the agent updates their beliefs about the latent system
of meaning using Bayes rule: \begin{equation}
\label{eq:joint_posterior}
\begin{array}{rcl}
P(\phi_k, \Theta | D_k)  & \propto &  P(D_k | \phi_k, \Theta) P(\phi_k, \Theta) \\
                           & =   & P(D_k | \phi_k) P(\phi_k | \Theta) P(\Theta)
\end{array}
\end{equation} This joint inference decomposes the learning problem into
two terms, a prior term \(P(\phi_k | \Theta)P(\Theta)\) and a likelihood
term \(P(D_k | \phi_k)\). The prior captures the idea that different
partners may share aspects of meaning in common. In the absence of
strong evidence that partner-specific language use departs from this
common structure, the agent ought to regularize toward background
knowledge of the population's conventions. The likelihood represents
predictions about how a particular partner will use language under
different systems of meaning.

The joint posterior over meanings in Eq. \ref{eq:joint_posterior} has
two consequences for convention formation. First, it allows agents to
maintain idiosyncratic partner-specific expectations \(\phi_k\) by
marginalizing over community-level uncertainty: \begin{equation}
P(\phi_k | D_k) = \int_{\Theta}P(D_k | \phi_k) P(\phi_k | \Theta) P(\Theta)  d\Theta
\end{equation} Second, the hierarchical structure provides an inductive
pathway for partner-specific data to inform beliefs about community-wide
conventions. Agents update their beliefs about \(\Theta\) by
marginalizing over data accumulated from different partners:
\begin{equation}
P(\Theta | D) = P(\Theta) \int_{\phi} P(D_k | \phi_k) P(\phi_k | \Theta) d\phi
\end{equation} where \(D = \bigcup_{k=1}^N D_k\),
\(\phi = \phi_1 \times \dots \times \phi_N\), and \(N\) is the number of
partners previous encountered.

After multiple partners are inferred to have a similar system of
meaning, beliefs about \(\Theta\) shift to represent this abstracted
knowledge: it becomes more likely that a novel partner will share it as
well. This hierarchical inference is sometimes referred to as ``sharing
of strength'' or ``partial pooling'' (Gelman \& Hill, 2006) because it
smoothly integrates aggregated data (as in complete-pooling models) with
domain-specific knowledge (as in no-pooling models).

\hypertarget{model-simulations}{%
\subsection{Model simulations}\label{model-simulations}}

\begin{CodeChunk}
\begin{figure*}[t!]

{\centering \includegraphics[width=400px]{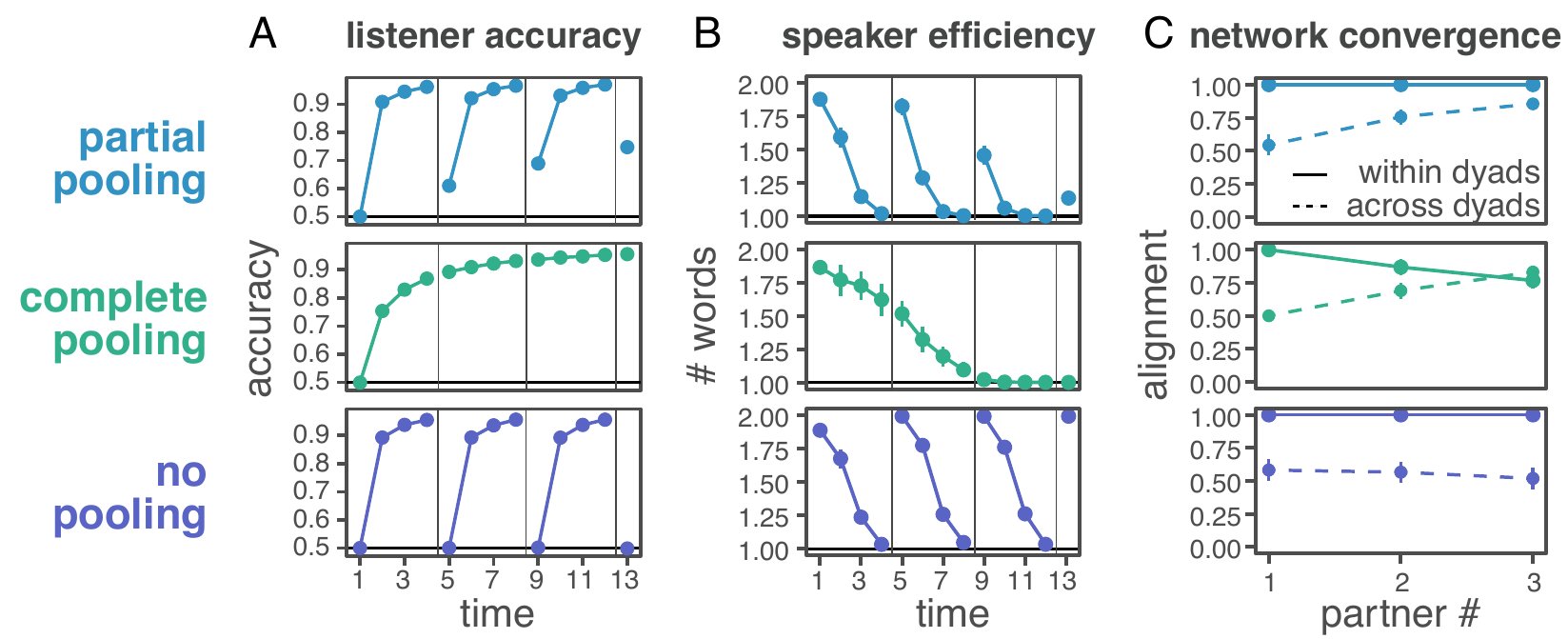} 

}

\caption[Model predictions for (A) listener accuracy, (B) speaker efficiency, and (C) network convergence across three partners]{Model predictions for (A) listener accuracy, (B) speaker efficiency, and (C) network convergence across three partners. Vertical lines indicate partner boundaries, and error bars are bootstrapped 95\% CIs across simulation runs.}\label{fig:model_results}
\end{figure*}
\end{CodeChunk}

We investigate the predictions of our partial-pooling model under three
increasingly complex scenarios, and compare them against the predictions
of complete-pooling and no-pooling models. In all of these scenarios,
speaker and listener agents play a reference game with a set of two
objects \(\mathcal{O} = \{o_1, o_2\}\). On each trial, one of these
objects is privately identified to the speaker as the \emph{target}.
They must then select from a set of utterances
\(\mathcal{U} = \{u_0, \dots, u_j\}\) to convey the identity of the
target to the listener. Upon hearing this utterance, the listener must
select which of the objects they believe to be the target. Both agents
then receive feedback. The resulting data \(D_k\) from an interaction
with partner \(k\) thus consists of utterance-object pairs
\(\{(u, o)_t\}\) for each trial \(t\).

In this reference game setting, we can explicitly specify the likelihood
and prior terms in Eq. (1). We consider a likelihood given by the
Rational Speech Act (RSA) framework, which formalizes the Gricean
assumption of cooperativity (Goodman \& Frank, 2016). A pragmatic
speaker \(S_1\) attempts to trade off informativity against the cost of
producing an utterance, while a pragmatic listener \(L_1\) inverts their
model of the speaker to infer the intended target. The chain of
recursive social reasoning grounds out in a \emph{literal listener}
\(L_0\), who identifies the intended target directly using a softmax
over the parameterized lexical meaning function
\(\mathcal{L}_{\phi_k}\). We assume, for simplicity, that \(\phi_k\) is
a \(|\mathcal{O}| \times |\mathcal{U}|\) real-valued matrix with entries
for each utterance-object pair, and \(\mathcal{L}_{\phi_k}(o,u)\) simply
looks up the entry for \((o,u)\). This model can be formally specified
as follows: \[
\begin{array}{rcl}
L_0(o | u, \phi_k) &\propto  & \exp\{\mathcal{L}_{\phi_k}(u,o)\} \\
S_1(u | o, \phi_k) &\propto &  \exp\{w_I \cdot \log L_0(o | u, \phi_k) - w_C \cdot c(u)\}   \\
L_1(o | u, \phi_k) &\propto  & S_1(u | o, \phi_k) P(o) 
\end{array}
\] \(c(u)\) is the cost of producing \(u\) and \(w_I\) and \(w_C\) are
free parameters controlling the relative weights on the informativity
and parsimony, respectively\footnote{Throughout our simulations, we set
  \(w_I = 11,~w_C = 7\). A grid search over parameter space revealed
  other regimes of behavior, but we leave broader exploration of this
  space to future work.}. Note that under each value of \(\phi_k\), the
\(S_1\) and \(L_1\) functions assign a probability to each word or
object that one's partner has chosen, thus yielding the likelihood of
the full set of observations \(P(D_k | \phi_k)\). In addition to using
the RSA likelihood for updating an agent's beliefs about \(\Theta\) and
\(\phi_k\), we can use the same functions to simulate the choices of
speaker and listener agents on a particular trial. That is, we sample
from the posterior predictive, marginalizing over the agent's current
beliefs about \(\phi_k\): \begin{align}
L(o|u) &\propto   \textstyle{\int_{\phi_k}} P(\phi_k | D_k) S_1(u|o, \phi_k)d\phi_k\label{eq:marginalized}\\
S(u|o) &\propto  \exp\{ \textstyle{\int_{\phi_k}} P(\phi_k | D_k)  w_I \log L_1(o| u, \phi_k) - w_C c(u)d\phi_k\}\nonumber
\end{align} Finally, we must specify the form of the hierarchical
lexical prior and a method to perform inference in this model. The
hyper-prior for \(\Theta\) is given by independent Gaussian
distributions for each matrix entry \(\Theta_{ij} \in \Theta\). We then
center the partner-specific prior \(\phi_{ij} \in \phi\) at the
corresponding value \(\Theta_{ij}\): \[\begin{array}{rcl}
P(\Theta_{ij}) & \sim & \mathcal{N}(0, 1)\\
P(\phi_{ij} | \Theta_{ij}) & \sim & \mathcal{N}(\Theta_{ij}, 1)
\end{array}\] These priors represent assumptions about how far
partner-specific learning can drift from the community-wide value.

For all simulations, we used variational inference as implemented in
WebPPL (Goodman \& Stuhlmüller, n.d.). Variational methods transform
probabilistic inference problems into optimization problems by
approximating the true posterior with a parameterized family.
Specifically, we make a \emph{mean-field} approximation and assume that
the full posterior can be factored into independent Gaussians for each
random variable. We then optimize the parameters of these posterior
Gaussians by minimizing the evidence lower bound (ELBO) objective (see
Murphy, 2012). On each trial, we run 50,000 gradient steps on previous
observations to obtain a posterior (Eq.~1) and compute the agent's
marginal prediction for the next observation by taking the expectation
over 50,000 samples from the variational guide (Eq. 4).\footnote{These
  details were chosen to ensure high-quality estimates of the model's
  behavior; see Discussion for other possible algorithms.}

\hypertarget{simulation-1-listener-accuracy-across-partners}{%
\subsubsection{Simulation 1: Listener accuracy across
partners}\label{simulation-1-listener-accuracy-across-partners}}

The key predictions of our partial-pooling model concern the pattern of
generalization across partners. In our first simulation, we consider the
partner-specificity of a \emph{listener}'s expectations about which
object is being referred to. To observe the model's behavior in the
simplest possible case, we assume the speaker has a vocabulary of two
single-word utterances \(\{u_1, u_2\}\) with equal cost and produces the
same utterance for the same target object (\(\{o_1, u_1\}\)) on every
trial. We introduce a new partner every 4 trials.

The probability the listener assigns to the target on each trial is
shown for the different models in Fig. \ref{fig:model_results}A. Under
the partial-pooling model (top row), the listener agent begins at chance
due to its uninformative prior, but after observing several trials of
evidence from the same partner, it rapidly infers the meaning of \(u_1\)
and learns to choose the true target with higher accuracy. When a second
partner is introduced, the agent's expectations revert nearly to their
original state, unlike a complete-pooling model (middle row). This
reversion is due to ambiguity about whether the behavior of the first
partner was idiosyncratic or attributable to community-level
conventions. In the absence of data from other partners, its
observations are more parsimoniously explained at the partner-specific
level. After observing multiple partners use \(u_1\) similarly, however,
we find that this knowledge has gradually been incorporated into
community-level expectations. This is evident in much stronger initial
expectations when introduced to the fourth partner (\(\sim\) 75\%
accuracy vs.~50\% with the first partner), unlike a no-pooling model
(bottom row).

\begin{CodeChunk}
\begin{figure*}[h]

{\centering \includegraphics[width=470px]{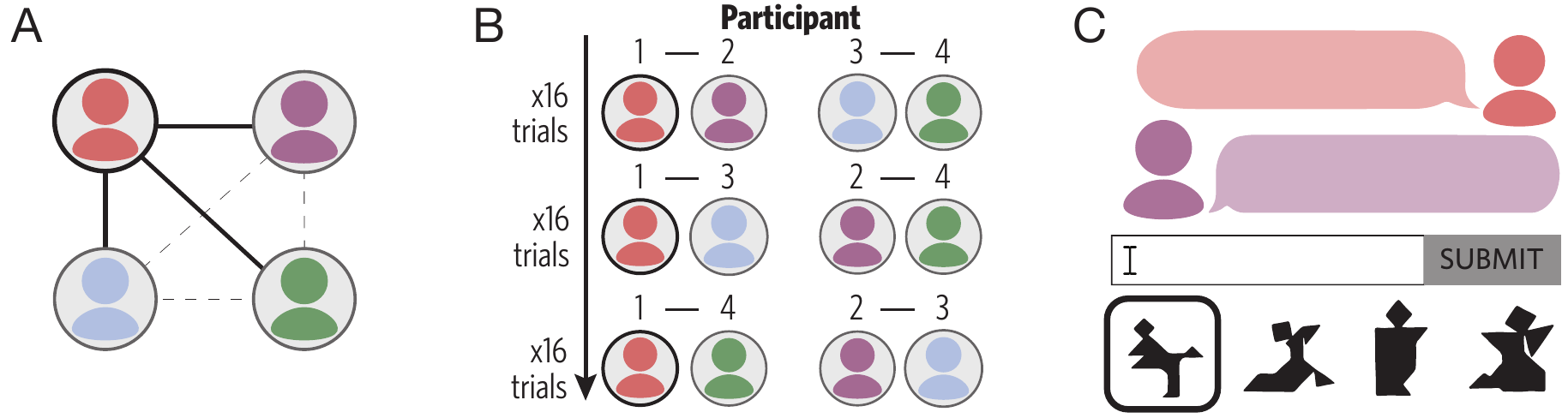} 

}

\caption[In our experiment, (A) participants were placed in fully-connected networks of 4, (B) paired in a round-robin schedule  with each neighbor, and (C) played a series of repeated reference games using tangram stimuli]{In our experiment, (A) participants were placed in fully-connected networks of 4, (B) paired in a round-robin schedule  with each neighbor, and (C) played a series of repeated reference games using tangram stimuli.}\label{fig:task1_display}
\end{figure*}
\end{CodeChunk}

\hypertarget{simulation-2-speaker-utterance-length-across-partners}{%
\subsubsection{Simulation 2: Speaker utterance length across
partners}\label{simulation-2-speaker-utterance-length-across-partners}}

Next, we examined our model's predictions about how a \emph{speaker}'s
referring expressions will change with successive listeners. While it
has been frequently observed that messages reduce in length across
repetitions with a single partner (Krauss \& Weinheimer, 1964) and
sharply revert back to longer utterances when a new partner is
introduced (Wilkes-Gibbs \& Clark, 1992), the key prediction
distinguishing our model concerns behavior across subsequent partner
boundaries. Complete-pooling accounts predict no reversion in number of
words when a new partner is introduced (Fig. \ref{fig:model_results}B,
middle row). No-pooling accounts predict that roughly the same initial
description length will re-occur with every subsequent interlocutor
(Fig. \ref{fig:model_results}B, bottom row). Here we show that a partial
pooling account predicts a more complex pattern of generalization.

We allowed a set of four primitive utterances,
\(\{u_1, u_2, u_3, u_4\}\), to be combined into conjunctions, e.g.
\(\{u_1+u_2, u_3+u_4\}\), which are assumed to have twice the utterance
cost. The meanings of these conjunctions were determined compositionally
from the values of the primitive utterances. We used a standard product
T-norm for conjunction. Because our values come from a Gaussian prior
and the T-norm is defined over \([0,1]\), we used logistic and logit
function to map values to the unit interval and back. Speakers do not
typically begin at chance over their \emph{entire} vocabulary, so we
introduced a weakly biased prior for \(\Theta\): two of the primitive
utterances initially applied more strongly to \(o_1\) and the other two
more strongly to \(o_2\). This weak bias leads to a preference for
conjunctions at the outset and thus allows us examine reduction.

We paired the speaker model with a fixed listener who always selected
the target, and ran 48 independent simulations. First, we found that
descriptions become more efficient over interaction with a single
partner: the model becomes more confident that shorter utterances will
be meaningful, so the marginal informativity provided by the conjunction
is not worth the additional cost (see Hawkins et al., 2017). Second, we
find that the speaker model reverts back to a longer description at the
first partner swap: evidence from one partner is relatively
uninformative about the community. Third, after interacting with several
partners, the model becomes more confident that one of the short labels
is shared across the entire community, and is correspondingly more
likely to begin a new interaction with it (Fig.
\ref{fig:model_results}B, top row).

\hypertarget{simulation-3-network-convergence}{%
\subsubsection{Simulation 3: Network
convergence}\label{simulation-3-network-convergence}}

The first two simulations presented a single adaptive agent with a fixed
partner to understand its gradient of generalization. In our final
simulation, we test the consequences of the proposed hierarchical
inference scheme for a network of \emph{interacting} agents. From each
individual agent's perspective, this simulation is identical to the
earlier ones (i.e.~a sequence of 3 different partners). Because all
agents are simultaneously making inferences about the others, however,
the network as a whole faces a coordination problem. For example, in the
first block, agents 1 and 2 may coordinate on using \(u_1\) to refer to
\(o_1\) while agent 3 and 4 coordinate on using \(u_2\). Once they swap
partners, they must negotiate this potential mismatch in usage. How does
the network as a whole manage to coordinate?

We used a round-robin scheme to schedule four agents into three blocks
of interaction, with agents taking turns in the speaker and listener
roles, and simulated 48 networks. We measured alignment at the
interaction-level by computing whether different agents produced the
same one-word utterances. We compared the alignment between currently
interacting agents (i.e. \emph{within} a dyad) to those who were not
interacting (i.e. \emph{across} dyads). Alignment across dyads was
initially near chance, reflecting the arbitrariness of whether speakers
reduce to \(u_1\) or \(u_2\). Under a no-pooling model (Fig.
\ref{fig:model_results}C, bottom row), subsequent blocks remain at
chance, as conventions need to be re-negotiated from scratch. Under a
complete-pooling model (Fig. \ref{fig:model_results}C, middle row),
agents persist with mis-calibrated expectations learned from previous
partners rather than adapting to their new partner, and
\emph{within-dyad} alignment deteriorates. By contrast, under our
partial-pooling model, alignment across dyads increases, suggesting that
hierarchical inference leads to emergent consensus (Fig.
\ref{fig:model_results}C, top row).

\hypertarget{behavioral-experiment}{%
\section{Behavioral experiment}\label{behavioral-experiment}}

\begin{CodeChunk}
\begin{figure*}[t!]

{\centering \includegraphics[width=370px]{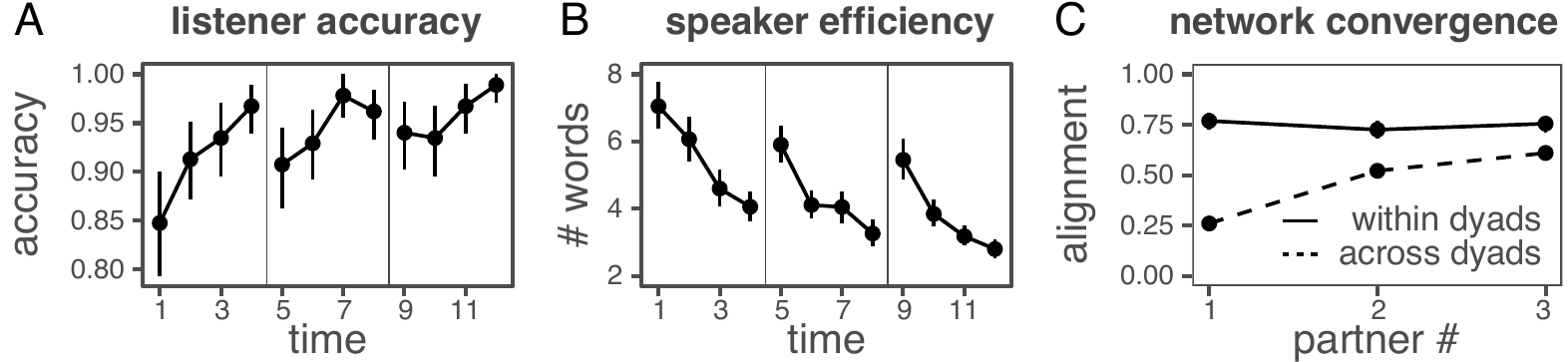} 

}

\caption[Results from behavioral experiment]{Results from behavioral experiment: (A) listener accuracy, (B) speaker efficiency, and (C) network convergence.}\label{fig:results}
\end{figure*}
\end{CodeChunk}

To evaluate the predictions observed in our simulations, we designed a
natural-language communication experiment following roughly the same
network design. Instead of anonymizing partners, as in many previous
empirical studies of convention formation, we divided the experiment
into blocks of extended dyadic interactions with stable, identifiable
partners (see Fay, Garrod, Roberts, \& Swoboda, 2010; Garrod \& Doherty,
1994 for similar designs). Each block was a full repeated reference
game, where participants had to coordinate on \emph{ad hoc} conventions
for how to refer to novel objects with their partner (Brennan \& Clark,
1996). Our partial-pooling model predicts that these conventions will
partially reset at partner boundaries, but agents should be increasingly
willing to transfer expectations from one partner to another.

\hypertarget{participants}{%
\subsubsection{Participants}\label{participants}}

We recruited 92 participants from Amazon Mechanical Turk to play a
series of interactive, natural-language reference games. Base pay was
set to \$3.00, with a 4 cent performance bonus for each correct
response.

\hypertarget{stimuli-and-procedure}{%
\subsubsection{Stimuli and procedure}\label{stimuli-and-procedure}}

Each participant was randomly assigned to one of 23 fully-connected
networks with three other participants as their `neighbors' (Fig.
\ref{fig:task1_display}A). Each network was then randomly assigned one
of three distinct ``contexts'' containing abstract tangram stimuli taken
from Clark and Wilkes-Gibbs (1986). The experiment was structured into a
series of three repeated reference games with different partners, using
these same four stimuli as referents. Partner pairings were determined
by a round-robin schedule (Fig. \ref{fig:task1_display}B). The trial
sequence for each reference game was composed of four repetition blocks,
where each target appeared once per block. Participants were randomly
assigned to speaker and listener roles and swapped roles on each block.
After completing sixteen trials with one partner, participants were
introduced to their next partner and asked to play the game again. This
process repeated until each participant had partnered with all three
neighbors. Because some pairs within the network took longer than
others, we sent participants to a temporary waiting room if their next
partner was not ready.

\vspace{1em}

Each trial proceeded as follows. First, one of the four tangrams in the
context was highlighted as the \emph{target object} for the speaker.
They were instructed to use a chatbox to communicate the identity of
this object to their partner, the listener (see Fig.
\ref{fig:task1_display}C). The two participants could engage freely in
dialogue through the chatbox but the listener must ultimately make a
selection from the array. Finally, both participants in a pair were
given full feedback on each trial about their partner's choice and
received bonus payment for each correct response. The order of the
stimuli on the screen was randomized on every trial to prevent the use
of spatial cues (e.g. `the one on the left'). The display also contained
an avatar representing their current partner to emphasize that they were
speaking to the same partner for an extended period.

\hypertarget{results}{%
\subsection{Results}\label{results}}

\vspace{-1em}

We evaluated participants' generalization behavior on the same three
metrics we used in our simulations: accuracy, utterance length, and
network convergence.

\hypertarget{listener-accuracy}{%
\subsubsection{Listener accuracy}\label{listener-accuracy}}

We first examined changes in the proportion of correct listener
selections. In particular, our partial pooling model predicts (1) gains
in accuracy within each partner and (2) drops in accuracy at partner
boundaries, but (3) overall improvement in initial interactions with
successive partners. To test the first prediction, we constructed a
logistic mixed-effects regression predicting trial-level listener
responses. We included a fixed effect of repetition block within partner
(1, 2, 3, 4), along with random intercepts and slopes for each
participant and each tangram. We found that accuracy improved over
successive repetitions with every partner,
\(b =0.69,\,~z = 3.87,\,~p<0.001\).

To test changes at partner boundaries, we constructed another regression
model. We coded the repetition blocks immediately before and after each
partner swap, and included this as a categorical fixed effect. Because
partner roles were randomized for each game, the same participant often
did not serve as listener in both blocks, so in addition to
tangram-level intercepts, we included random slopes and intercepts at
the \emph{network} level (instead of the participant level). We found
that across the two partner swaps, accuracy dropped significantly,
\(b=-1.56,\,~z=-2,\,~p<0.05\), reflecting partner-specificity of
meaning. Finally, to test whether performance improves for the
\emph{very first} interaction with each new partner, before observing
any partner-specific information, we examined the simple effect of
partner number on the trials immediately after the partner swap
(\(t=\{1,5,9\}\)). As predicted, we found a significant improvement in
performance, \(b=0.57,\,~z=2.72,\,~p<0.01\), suggesting that listeners
are bringing increasingly well-calibrated expectations into interactions
with novel neighbors (see Fig. \ref{fig:results}A).

\hypertarget{speaker-utterance-length}{%
\subsubsection{Speaker utterance
length}\label{speaker-utterance-length}}

Next, as a measure of coding efficiency, we calculated the raw length
(in words) of the utterance produced on each trial. We then tested
analogues of the same three predictions we tested in the previous
section using the same mixed-effects models, but using a linear
regression on the continuous measure of efficiency instead of accuracy
(see Fig. \ref{fig:results}B). We log-transformed utterance lengths for
stability. We found that speakers reduced utterance length with every
partner, \(b=-0.19,\,~t(34)=-9.88,~p<0.001\), increased length across
partner-boundaries, \(b=0.43,\,~t(22)=4.4,\,~p<0.001\), and decreased
the length of their \emph{initial descriptions} as they interacted with
more partners on their network, \(b =-0.2\), \(t(516.5)=-6.07\),
\(p<0.001\) (see Fig. \ref{fig:results}B).

\hypertarget{network-convergence}{%
\subsubsection{Network convergence}\label{network-convergence}}

In this section, we examine the actual \emph{content} of pacts and test
whether these coarse signatures of generalization actually lead to
increased alignment across the network, as predicted. Specifically, we
extend the `exact matching' measure of alignment used in Simulation 3 to
natural language production by examining whether the \emph{intersection}
of words produced by different speakers was non-empty. We excluded a
list of common stop words (e.g. `the', `both') to focus on the core
conceptual content of pacts; using the size of the intersection instead
of the binary variable yielded similar results.

As in our simulation, the main comparison of interest was between
currently interacting participants and participants who are not
interacting: we predicted that within-pair alignment should stay
consistently high while (tacit) alignment between non-interacting pairs
will increase. We thus constructed a mixed-effects logistic regression
including fixed effects of pair type (within vs.~across), partner
number, and their interaction. We included random intercepts at the
tangram level and maximal random effects at the network level
(i.e.~intercept, both main effects, and the interaction). As predicted,
we found a significant interaction (\(b=-0.85,~z=-5.69,\,~p<0.001\); see
Fig. \ref{fig:results}C). Although different pairs in a network may
initially use different labels, these labels begin to align over
subsequent interactions.

\hypertarget{discussion}{%
\section{Discussion}\label{discussion}}

How do community-level conventions emerge from local interactions? In
this paper, we proposed a partial-pooling account, formalized as a
hierarchical Bayesian model, where conventions represent the shared
structure that agents ``abstract away'' from partner-specific
interactions. Unlike complete-pooling accounts, this model allows for
partner-specific common ground to override community-wide expectations
given sufficient experience with a partner, or in the absence of strong
conventions. Unlike no-pooling accounts, it allows networks to converge
on more efficient and accurate expectations about novel partners. We
conducted a series of simulations investigating the model's
generalization behavior, and evaluated these predictions with a
natural-language behavioral experiment on small networks.

Hierarchical Bayesian models have several other properties of
theoretical interest for convention formation. First, they offer a
``blessing of abstraction'' (Goodman et al., 2011), where
community-level conventions may be learned even with relatively sparse
input from each partner, as long as there is not substantial variance in
the population. Second, they are more robust to partner-specific
deviations from conventions (e.g.~interactions with children or
non-native speakers) than complete-pooling models relying on a fixed set
of memory slots or a single mental ``inventory.'' This robustness is due
to their ability to ``explain away'' outliers without community-level
expectations being affected.

While our behavioral data is inconsistent with complete-pooling and
no-pooling accounts, there remain deep theoretical connections between
our hierarchical Bayesian formulation and alternative theories of
generalization (e.g. Rogers \& McClelland, 2004; Doumas, Hummel, \&
Sandhofer, 2008; Marcus, 2001). For example, recent neural network
algorithms such as Model-Agnostic Meta-Learning (MAML; Finn, Abbeel, \&
Levine, 2017), which attempt to learn general parameter settings
(e.g.~conventions) that can rapidly adapt to different specific tasks
(e.g.~partners), have been shown to be equivalent to hierarchical Bayes
under certain conditions (Grant, Finn, Levine, Darrell, \& Griffiths,
2018). Such neural network instantiations may scale better in practice
to natural language in more complex referential spaces, and may only
require a handful of gradient steps to successfully adapt (Hawkins et
al., 2019b). Alternatively, if different partners are represented as
different ``contexts'' (Brown-Schmidt, Yoon, \& Ryskin, 2015), then
context-dependent reinforcement learning mechanisms may produce similar
predictions (Gershman \& Niv, 2015). These theoretical connections
expose a common underlying view that conventions emerge from a group of
agents discovering latent structure in coordination problems by adapting
to each idiosyncratic partner along the way.

Real-world communities are much more complex than the simple networks we
considered: each speaker takes part in a number of overlapping
subcommunities. For example, we use partially distinct conventions
depending on whether we are communicating with psychologists, friends
from high school, bilinguals, or children. For future work to address
the full scale of an individual's network of communities, additional
social knowledge about these communities must be learned and represented
in the generative model (e.g.~Gershman et al, 2017). Additionally, the
number of distinct speakers an individual is exposed to (Lev-Ari, 2018)
and the connectivity dynamics of their community (Segovia-Martín,
Walker, Fay, \& Tamariz, 2020) remain key manipulations to explore. Our
results are a promising first step, providing evidence that hierarchical
generalization from partner-specific representations may be a
foundational cognitive mechanism for establishing conventionality at the
group level.

\hypertarget{acknowledgments}{%
\section{Acknowledgments}\label{acknowledgments}}

\small

We thank Bill Thompson, Judith Fan, and Kenny Smith for insightful
conversations. This work was supported by NSF SPRF-FR grant \#1911835 to
RDH, TLG, and AEG, and CV Starr Fellowship to RDH.

\vspace{2em}
\fbox{\parbox[b][][c]{7.3cm}{\centering {All code and materials for simulations, data analyses, and web experiment available at: \\
\href{https://github.com/hawkrobe/conventions_model}{\url{https://github.com/hawkrobe/conventions_model}}
}}}
\vspace{2em}

\noindent

\normalsize

\hypertarget{references}{%
\section{References}\label{references}}

\setlength{\parindent}{-0.1in} 
\setlength{\leftskip}{0.125in}

\noindent

\hypertarget{refs}{}
\leavevmode\hypertarget{ref-barr_establishing_2004}{}%
Barr, D. J. (2004). Establishing conventional communication systems: Is
common knowledge necessary? \emph{Cognitive Science}, \emph{28}(6),
937--962.

\leavevmode\hypertarget{ref-bergen_pragmatic_2016}{}%
Bergen, L., Levy, R., \& Goodman, N. (2016). Pragmatic reasoning through
semantic inference. \emph{Semantics and Pragmatics}, \emph{9}(20).

\leavevmode\hypertarget{ref-BrennanClark96_ConceptualPactsConversation}{}%
Brennan, S., \& Clark, H. (1996). Conceptual pacts and lexical choice in
conversation. \emph{Journal of Experimental Psychology: Learning,
Memory, and Cognition}, \emph{22}(6), 1482--1493.

\leavevmode\hypertarget{ref-brown2015people}{}%
Brown-Schmidt, S., Yoon, S. O., \& Ryskin, R. A. (2015). People as
contexts in conversation. In \emph{Psychology of learning and
motivation} (Vol. 62, pp. 59--99). Elsevier.

\leavevmode\hypertarget{ref-centola_spontaneous_2015}{}%
Centola, D., \& Baronchelli, A. (2015). The spontaneous emergence of
conventions: An experimental study of cultural evolution.
\emph{Proceedings of the National Academy of Sciences}, \emph{112}(7),
1989--1994.

\leavevmode\hypertarget{ref-clark_using_1996}{}%
Clark, H. H. (1996). \emph{Using language}. Cambridge, England:
Cambridge University Press.

\leavevmode\hypertarget{ref-clark_referring_1986}{}%
Clark, H. H., \& Wilkes-Gibbs, D. (1986). Referring as a collaborative
process. \emph{Cognition}, \emph{22}(1), 1--39.

\leavevmode\hypertarget{ref-davidson_nice_1986}{}%
Davidson, D. (1986). A nice derangement of epitaphs. \emph{Philosophical
Grounds of Rationality: Intentions, Categories, Ends}, \emph{4},
157--174.

\leavevmode\hypertarget{ref-doumas2008theory}{}%
Doumas, L. A., Hummel, J. E., \& Sandhofer, C. M. (2008). A theory of
the discovery and predication of relational concepts.
\emph{Psychological Review}, \emph{115}(1), 1.

\leavevmode\hypertarget{ref-eckert_three_2012}{}%
Eckert, P. (2012). Three waves of variation study: The emergence of
meaning in the study of sociolinguistic variation. \emph{Annual Review
of Anthropology}, \emph{41}, 87--100.

\leavevmode\hypertarget{ref-fay_interactive_2010}{}%
Fay, N., Garrod, S., Roberts, L., \& Swoboda, N. (2010). The interactive
evolution of human communication systems. \emph{Cognitive Science},
\emph{34}(3), 351--386.

\leavevmode\hypertarget{ref-FinnAbeelLevine17_MAML}{}%
Finn, C., Abbeel, P., \& Levine, S. (2017). Model-agnostic meta-learning
for fast adaptation of deep networks. \emph{arXiv Preprint
arXiv:1703.03400}.

\leavevmode\hypertarget{ref-garrod_conversation_1994}{}%
Garrod, S., \& Doherty, G. (1994). Conversation, co-ordination \&
convention: An empirical investigation of how groups establish
linguistic conventions. \emph{Cognition}, \emph{53}(3), 181--215.

\leavevmode\hypertarget{ref-gelman2006data}{}%
Gelman, A., \& Hill, J. (2006). \emph{Data analysis using regression and
multilevel/hierarchical models}. Cambridge, England: Cambridge
University Press.

\leavevmode\hypertarget{ref-gershman2015novelty}{}%
Gershman, S. J., \& Niv, Y. (2015). Novelty and inductive generalization
in human reinforcement learning. \emph{Topics in Cognitive Science},
\emph{7}(3), 391--415.

\leavevmode\hypertarget{ref-gershman_learning_2017}{}%
Gershman, S., Pouncy, H., \& Gweon, H. (2017). Learning the Structure of
Social Influence. \emph{Cognitive Science}, \emph{41}.

\leavevmode\hypertarget{ref-GoodmanFrank16_RSATiCS}{}%
Goodman, N. D., \& Frank, M. C. (2016). Pragmatic language
interpretation as probabilistic inference. \emph{Trends in Cognitive
Sciences}, \emph{20}(11), 818--829.

\leavevmode\hypertarget{ref-GoodmanStuhlmuller14_DIPPL}{}%
Goodman, N. D., \& Stuhlmüller, A. (n.d.). The design and implementation
of probabilistic programming languages. Retrieved from
\url{http://dippl.org}

\leavevmode\hypertarget{ref-GoodmanUllmanTenenbaum11_TheoryOfCausality}{}%
Goodman, N. D., Ullman, T. D., \& Tenenbaum, J. B. (2011). Learning a
theory of causality. \emph{Psychological Review}, \emph{118}(1),
110--119.

\leavevmode\hypertarget{ref-grant_recasting_2018}{}%
Grant, E., Finn, C., Levine, S., Darrell, T., \& Griffiths, T. (2018).
Recasting Gradient-Based Meta-Learning as Hierarchical Bayes.
\emph{arXiv Preprint arXiv:1801.08930}.

\leavevmode\hypertarget{ref-hawkins_convention-formation_2017}{}%
Hawkins, R. D., Frank, M. C., \& Goodman, N. D. (2017).
Convention-formation in iterated reference games. In \emph{Proceedings
of the 39th Annual Meeting of the Cognitive Science Society}.

\leavevmode\hypertarget{ref-hawkins2019emergence}{}%
Hawkins, R. D., Goodman, N. D., \& Goldstone, R. L. (2019a). The
emergence of social norms and conventions. \emph{Trends in Cognitive
Sciences}, \emph{23}(2), 158--169.

\leavevmode\hypertarget{ref-hawkins2019continual}{}%
Hawkins, R. D., Kwon, M., Sadigh, D., \& Goodman, N. D. (2019b).
Continual adaptation for efficient machine communication. \emph{arXiv
Preprint arXiv:1911.09896}.

\leavevmode\hypertarget{ref-KleinschmidtJaeger15_RobustSpeechPerception}{}%
Kleinschmidt, D., \& Jaeger, T. F. (2015). Robust speech perception:
Recognize the familiar, generalize to the similar, and adapt to the
novel. \emph{Psychological Review}, \emph{122}(2), 148--203.

\leavevmode\hypertarget{ref-krauss_changes_1964}{}%
Krauss, R. M., \& Weinheimer, S. (1964). Changes in reference phrases as
a function of frequency of usage in social interaction: A preliminary
study. \emph{Psychonomic Science}, \emph{1}(1-12), 113--114.

\leavevmode\hypertarget{ref-lev2018social}{}%
Lev-Ari, S. (2018). Social network size can influence linguistic
malleability and the propagation of linguistic change. \emph{Cognition},
\emph{176}, 31--39.

\leavevmode\hypertarget{ref-lewis_convention:_1969}{}%
Lewis, D. (1969). \emph{Convention: A philosophical study}. Harvard
University Press.

\leavevmode\hypertarget{ref-marcus2018algebraic}{}%
Marcus, G. F. (2001). \emph{The algebraic mind: Integrating
connectionism and cognitive science}. MIT press.

\leavevmode\hypertarget{ref-metzing_when_2003}{}%
Metzing, C., \& Brennan, S. E. (2003). When conceptual pacts are broken:
Partner-specific effects on the comprehension of referring expressions.
\emph{Journal of Memory and Language}, \emph{49}(2).

\leavevmode\hypertarget{ref-murphy2012machine}{}%
Murphy, K. P. (2012). \emph{Machine learning: A probabilistic
perspective}. MIT press.

\leavevmode\hypertarget{ref-rogers2004semantic}{}%
Rogers, T. T., \& McClelland, J. L. (2004). \emph{Semantic cognition: A
parallel distributed processing approach}. MIT press.

\leavevmode\hypertarget{ref-1f3d6484b7204ecf8ae961df38c17ef1}{}%
Segovia-Martín, J., Walker, B., Fay, N., \& Tamariz, M. (2020). Network
connectivity dynamics, cognitive biases and the evolution of cultural
diversity in round-robin interactive micro-societies. \emph{Cognitive
Science}.

\leavevmode\hypertarget{ref-steels_self-organizing_1995}{}%
Steels, L. (1995). A self-organizing spatial vocabulary.
\emph{Artificial Life}, \emph{2}(3), 319--332.

\leavevmode\hypertarget{ref-tenenbaum_how_2011}{}%
Tenenbaum, J. B., Kemp, C., Griffiths, T. L., \& Goodman, N. D. (2011).
How to grow a mind: Statistics, structure, and abstraction.
\emph{Science}, \emph{331}(6022), 1279--1285.

\leavevmode\hypertarget{ref-wilkes-gibbs_coordinating_1992}{}%
Wilkes-Gibbs, D., \& Clark, H. H. (1992). Coordinating beliefs in
conversation. \emph{Journal of Memory and Language}, \emph{31}(2),
183--194.

\leavevmode\hypertarget{ref-young_evolution_2015}{}%
Young, H. P. (2015). The Evolution of Social Norms. \emph{Annual Review
of Economics}, \emph{7}, 359--387.

\bibliographystyle{apacite}

\end{document}